\documentclass[runningheads]{llncs}

\usepackage[T1]{fontenc}
\usepackage{graphicx}
\usepackage{amsmath,amssymb,amsfonts}
\usepackage{booktabs}
\usepackage{multirow}
\usepackage{tabularx}
\usepackage{algorithm}
\usepackage{algpseudocode}
\usepackage{xcolor}
\usepackage{float}
\usepackage{pgfplots}
\usepackage{svg}
\usepackage{wrapfig}
\usepackage{enumitem}
\setlist{nosep}
\usepackage{subcaption}
\usepackage[colorlinks=true, citecolor=blue, linkcolor=blue, urlcolor=blue]{hyperref}

\usepackage{orcidlink}

\makeatletter
\newcommand{\printfnsymbol}[1]{%
  \textsuperscript{\@fnsymbol{#1}}%
}
\makeatother

\begin{document}
\title{Content-Preserving Principal Feature Blending for Training-Free Style Personalization}

\titlerunning{Content-Preserving Principal Feature Blending}

\author{
Le-Hoang Bui\inst{1,2}\orcidlink{0009-0003-1162-234X}\thanks{These authors contributed equally to this work.} \and
Hoang-Nguyen Cao\inst{1, 2}\orcidlink{0009-0003-0459-730X} \printfnsymbol{1}  \and
Trung-Nghia Le\inst{1,2}\orcidlink{0000-0002-7363-2610}\thanks{Corresponding author.} \and
Minh-Triet Tran\inst{1,2}\orcidlink{0000-0003-3046-3041} \and
Dinh-Khoi Vo\inst{1,2}\orcidlink{0000-0001-8831-8846} 
}

\authorrunning{Le-Hoang Bui et al.}

\institute{
University of Science, Ho Chi Minh City, Vietnam
\and
Vietnam National University, Ho Chi Minh City, Vietnam
\\
\email{
\{blhoang23, chnguyen23\}@apcs.fitus.edu.vn, \\
\{ltnghia, tmtriet\}@fit.hcmus.edu.vn, vdkhoi@selab.hcmus.edu.vn
}
}

\maketitle       

%

\begin{abstract}
Reference-based style personalization should reproduce the appearance of a style reference while preserving the subject specified by a content prompt. In visual autoregressive models, dominant reference features can also carry object and spatial information, causing content leakage. We propose Content-Preserving Principal Feature Blending (CoPFB), a training-free method that estimates a low-rank content basis from a clean generation branch and removes the content-aligned component of the style residual. CoPFB applies rank-one blending at the first two autoregressive scales, followed by weaker content-projected refinement at the next three scales. On 1,900 combinations of Parti content prompts and FineStyle reference images, CoPFB improves text alignment from 0.293 to 0.297 and the harmonic score from 0.391 to 0.393 over PFB+SAC, while style similarity decreases from 0.590 to 0.577. CoPFB requires no fine-tuning or learnable parameters. Project page: \url{https://leehoang2710.github.io/CoPFB/}.

\end{abstract}

\begin{figure}[t!]
    \centering
    \includegraphics[width=\textwidth]{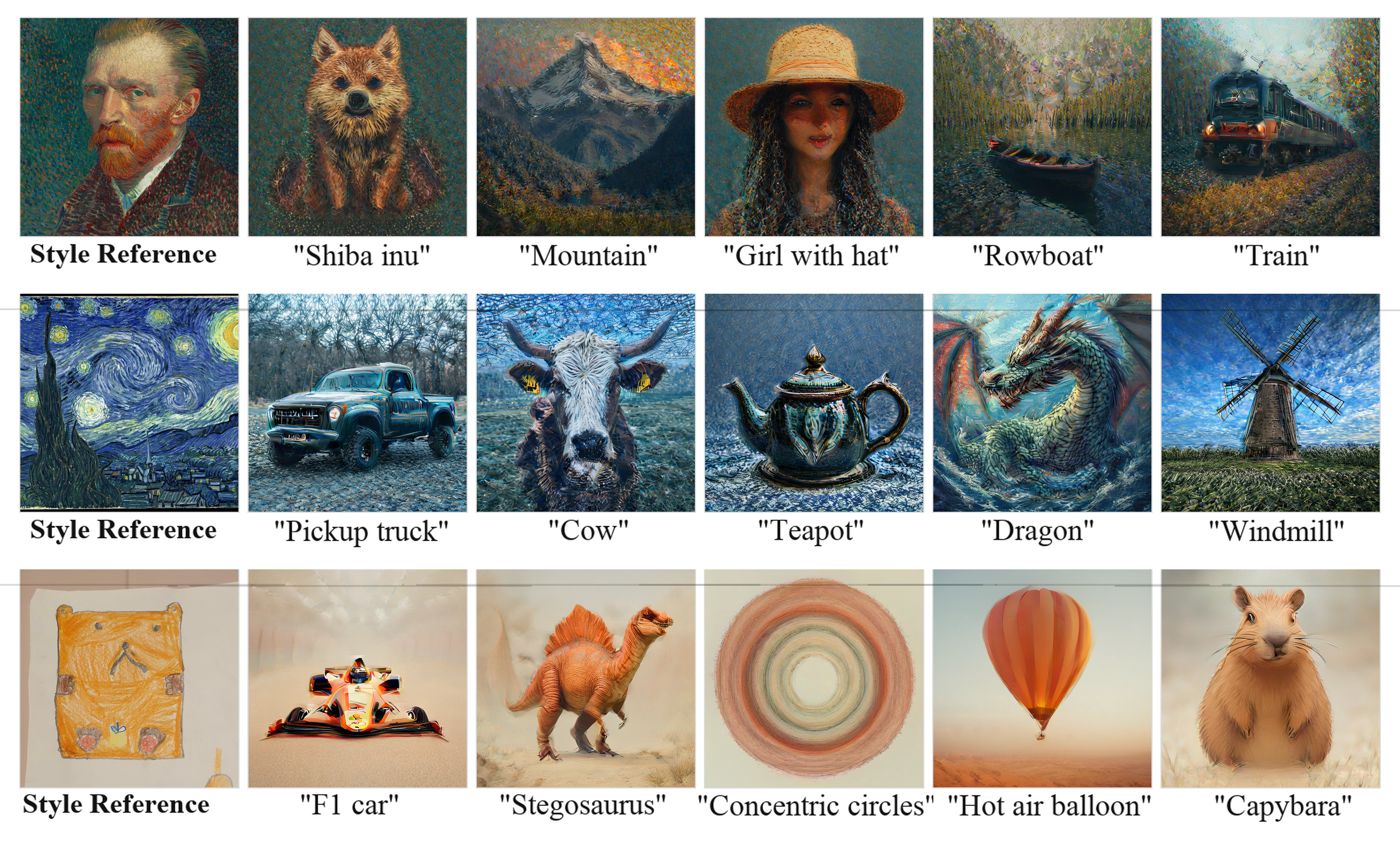}
    \caption{CoPFB results. Each row shows a style reference followed by outputs for different content prompts.}
    \label{fig:teaser}
\end{figure}

\section{Introduction}
Text-to-image generation has advanced through diffusion and visual autoregressive models. Visual Autoregressive Modeling (VAR)~\cite{tian2024visual} generates images through next-scale prediction, and Infinity~\cite{han2024infinity} extends this paradigm to high-resolution synthesis. This coarse-to-fine process provides a natural interface for transferring the appearance of a reference image while preserving content specified by a prompt.
The central challenge is content-style disentanglement: injecting reference features can strengthen color and texture transfer while also copying objects or spatial structure, thereby changing the requested subject, geometry, or composition.
Early style-transfer methods use per-image optimization~\cite{Gatys_2016_CVPR}, whereas feed-forward approaches learn one~\cite{johnson2016perceptual,ulyanov2016texture}, several~\cite{Chen_2017_CVPR,Dumoulin2017}, or arbitrary styles~\cite{huang2017arbitrary,li2017universal,park2019arbitrary}. In visual autoregressive models, CSD-VAR~\cite{nguyen2025csdvar} uses scale-aware optimization and SVD-based rectification. Lee et al.~\cite{Lee_2026_CVPR} introduce Principal Feature Blending (PFB), which injects dominant SVD components of scale-wise reference features, and Structural Attention Correction (SAC), which preserves prompt structure. Kingfisher~\cite{vo2026kingfisher} instead decomposes and recomposes early scale codes for content-style personalization. Because singular-value magnitude ranks feature energy rather than semantic role, the PFB residual can still overlap with content-related directions.

We study the role of each generation scale and the content sensitivity of dominant singular components. Based on these analyses, we propose \textbf{Content-Preserving Principal Feature Blending
(CoPFB)}, a training-free method to estimate a low-rank content basis from an unmodified generation branch and removes the component of the PFB residual~\cite{Lee_2026_CVPR} that lies in this basis. The filtered residual is injected into the editable branch and combined with SAC. 
We evaluate CoPFB on 1,900 content--style pairs built from Parti prompts~\cite{yu2022parti} and FineStyle references~\cite{zhang2024finestyle}, comparing our content-projection based method with PFB+SAC~\cite{Lee_2026_CVPR} and providing further ablations across different settings, all under identical generation settings. The proposed hybrid schedule improves prompt fidelity and achieves the best harmonic content--style score, while reducing inference time from approximately 3.8 to 2.0 seconds per image relative to PFB+SAC.

Our contributions are summarized as follows:
\begin{itemize}
    \item We show that principal-component-based style injection remains sensitive to reference content.
    \item We propose a training-free projection that suppresses content-aligned components of the PFB residual and a scale-dependent schedule that balances prompt and style alignment.
\end{itemize}

\section{Related Work}
\paragraph{\textbf{Text-to-Image and Visual Autoregressive Models.}}
Text-to-image systems use autoregressive~\cite{ramesh2021dalle}, diffusion~\cite{ho2020denoising}, and latent-diffusion~\cite{rombach2022high} formulations. VAR~\cite{tian2024visual} introduces next-scale prediction, which Infinity~\cite{han2024infinity} extends with bitwise autoregressive modeling. Its coarse-to-fine process exposes when composition, structure, and texture become controllable.
\paragraph{\textbf{Neural Style Transfer.}}

Gatys et al.~\cite{Gatys_2016_CVPR} formulate style transfer as feature matching, while feed-forward methods improve efficiency~\cite{johnson2016perceptual}. AdaIN~\cite{huang2017arbitrary}, WCT~\cite{li2017universal}, and attention-based methods~\cite{park2019arbitrary,deng2022stytr2} align feature statistics or content-style correspondences. CoPFB studies scale-wise visual autoregressive personalization.
\section{Proposed Method}
\begin{figure}[!t]
    \centering

    \begin{subfigure}[t]{\textwidth}
        \centering
        \includegraphics[
            width=\linewidth,
            height=0.20\textheight,
            keepaspectratio
        ]{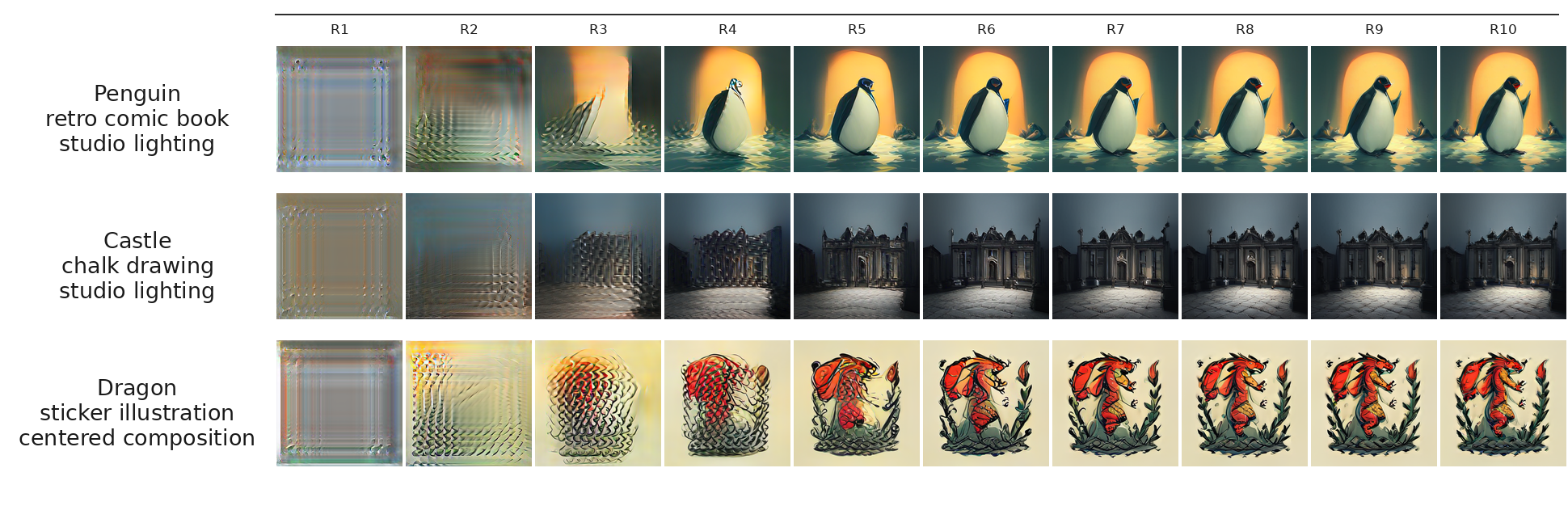}
        \caption{Scale-wise reconstruction from global composition to local detail.}
        \label{fig:stepwise_showcase}
    \end{subfigure}

    \vspace{-1mm}
 
    \begin{subfigure}[t]{\textwidth}
        \centering
        \includegraphics[
            width=\linewidth,
            height=0.20\textheight,
            keepaspectratio
        ]{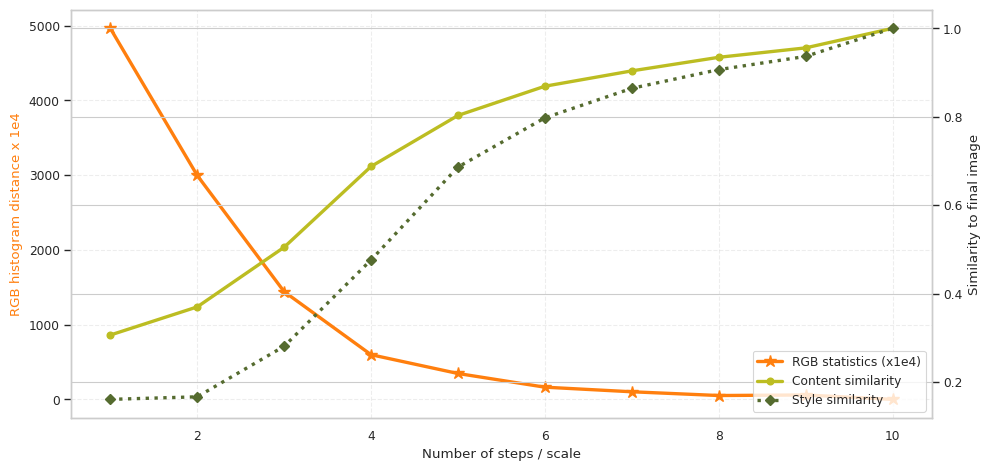}
        \caption{Dynamics over 50 prompts. RGB statistics change most at $R_1$--$R_2$; content and style similarity rise mainly at $R_3$--$R_5$.}
        \label{fig:step_dynamics}
    \end{subfigure}

    \caption{Coarse-to-fine dynamics of next-scale prediction.}
    \label{fig:study1_scale_dynamics}
\end{figure}

\subsection{Empirical Analysis}
\label{sec:analysis}

\paragraph{\textbf{Scale-wise generation and intervention.}}
\label{sec:next_scale_analysis}

Following Lee et al.~\cite{Lee_2026_CVPR}, we compare each scale reconstruction with the final output over 50 prompts using RGB-histogram distance, content similarity, and style similarity. We denote the ten resolutions by $R_s$, with zero-based index $s\in\{0,\ldots,9\}$. Fig.~\ref{fig:study1_scale_dynamics} shows that RGB statistics stabilize at $R_1$--$R_2$, similarities rise most at $R_3$--$R_5$, and later scales refine detail.
\begin{figure}[!t]
    \centering
    \vspace{0mm}
    \begin{subfigure}[t]{\textwidth}
        \centering
        \includegraphics[
            width=\linewidth,
            height=0.22\textheight,
            keepaspectratio
        ]{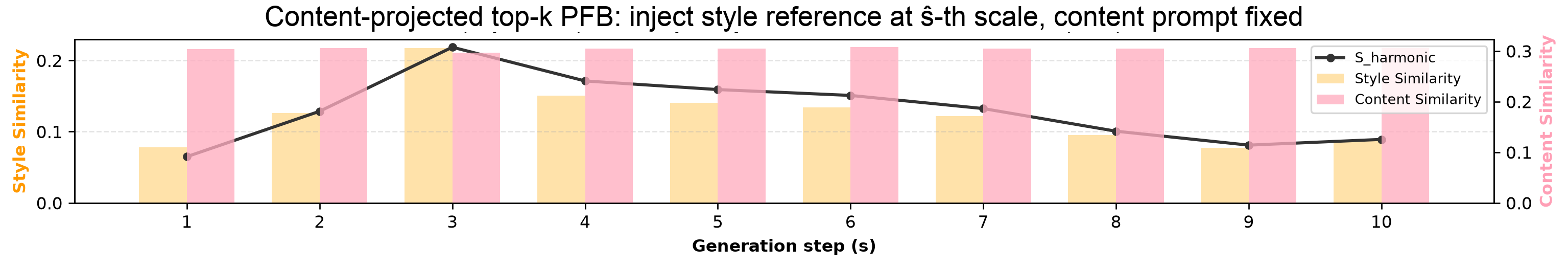}
        \caption{Step-wise injection scores peak at $R_3$.}
        \label{fig:svd_step_ablation}
    \end{subfigure}

    \vspace{-1mm}

    \begin{subfigure}[t]{\textwidth}
        \centering
        \includegraphics[
            width=\linewidth,
            height=0.22\textheight,
            keepaspectratio
        ]{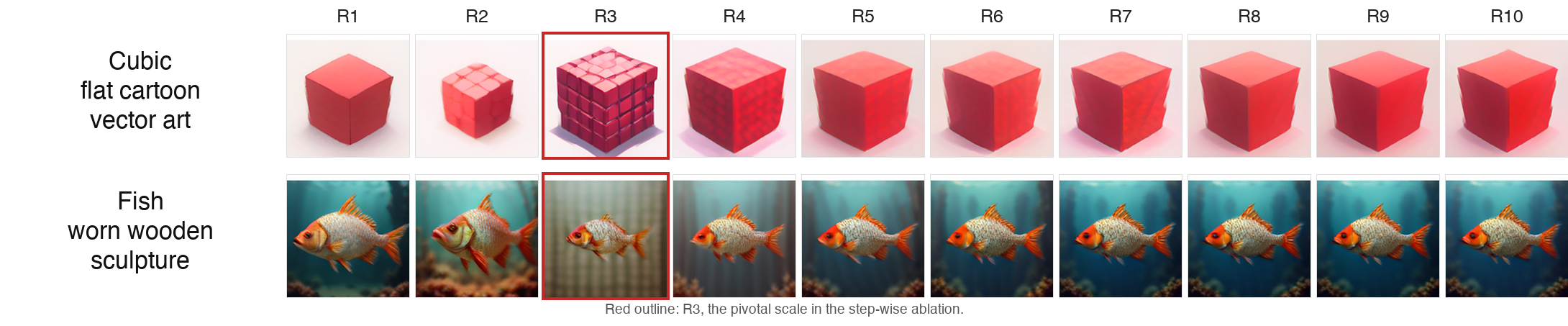}
        \caption{Changing the intervention scale affects style transfer and content preservation.}
        \label{fig:step3_examples}
    \end{subfigure}

    \caption{Quantitative and qualitative identification of the pivotal scale.}
    \label{fig:study2_pivotal_scale}
\end{figure}

\paragraph{Pivotal intervention scale.}
\label{sec:pivotal_feature_analysis}

We inject the same content-projected top-$k$ PFB residual at each Infinity-2B scale while fixing the prompt, reference, seed, and sampling configuration. Fig.~\ref{fig:study2_pivotal_scale} shows the highest harmonic score at $R_3$, motivating a strong intermediate intervention followed by attenuated updates.
\begin{figure*}[!t]
    \centering
    \captionsetup{skip=3pt}
    \begin{minipage}{\textwidth}
        \centering
        \includegraphics[width=\textwidth]{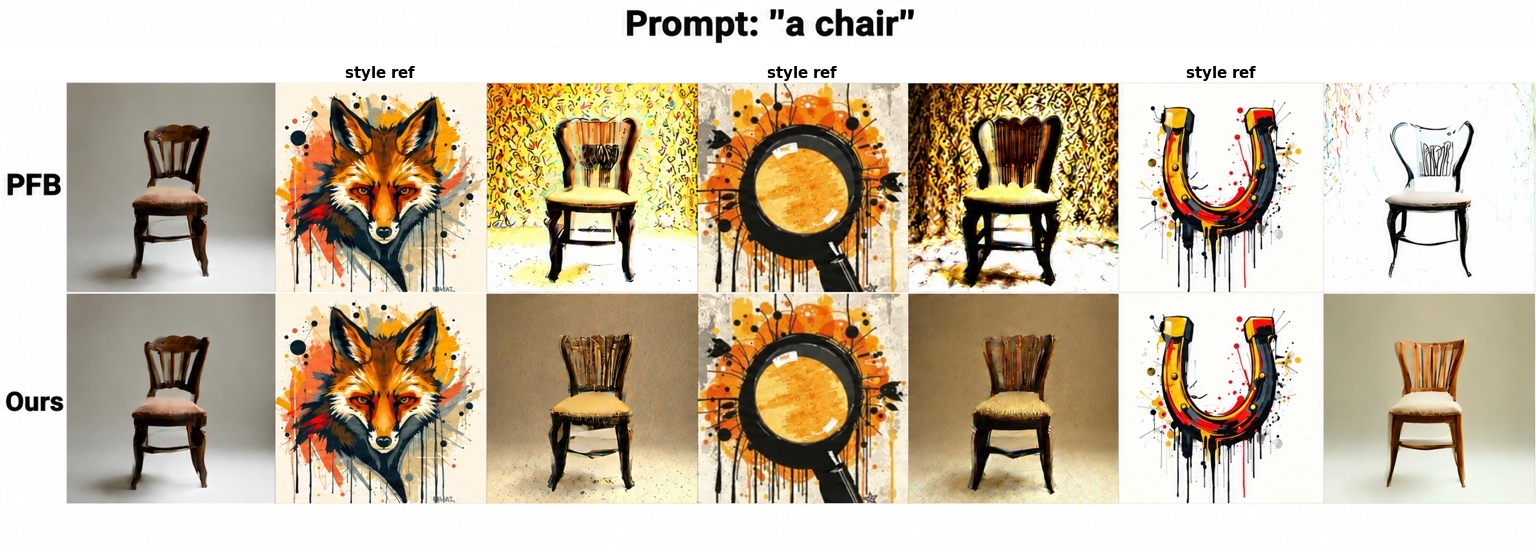}\par\vspace{-1.5mm}
        \includegraphics[width=\textwidth]{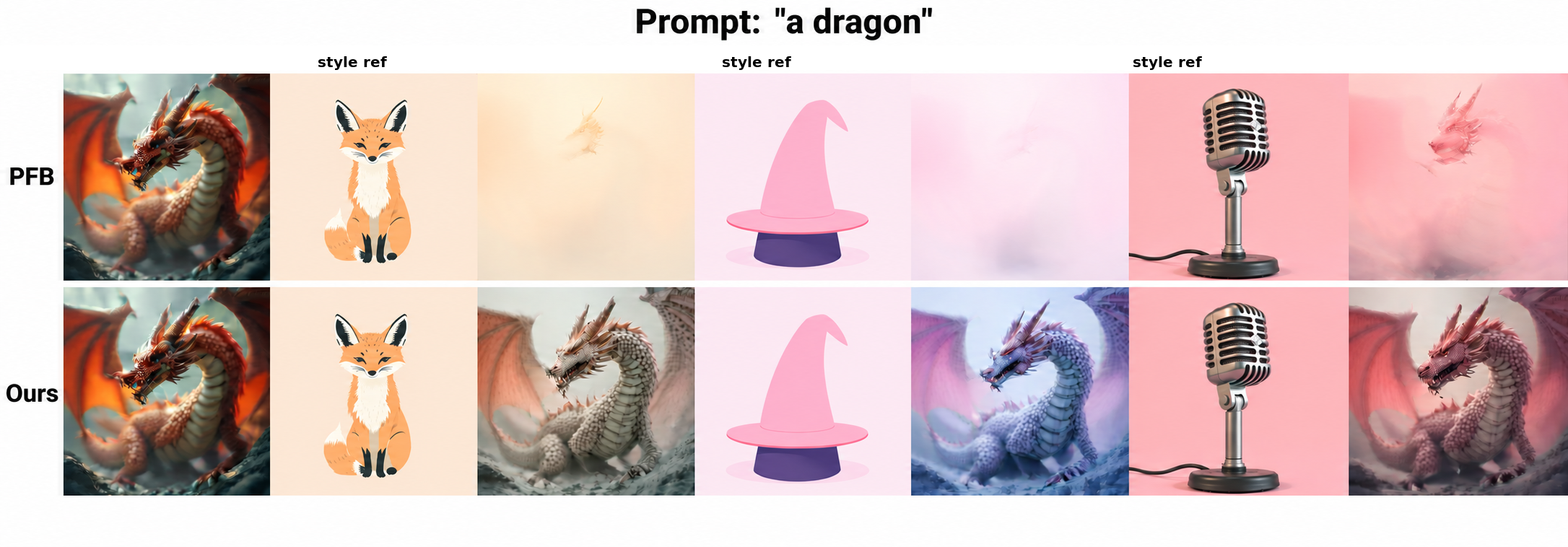}\par\vspace{-1.5mm}
        \includegraphics[width=\textwidth]{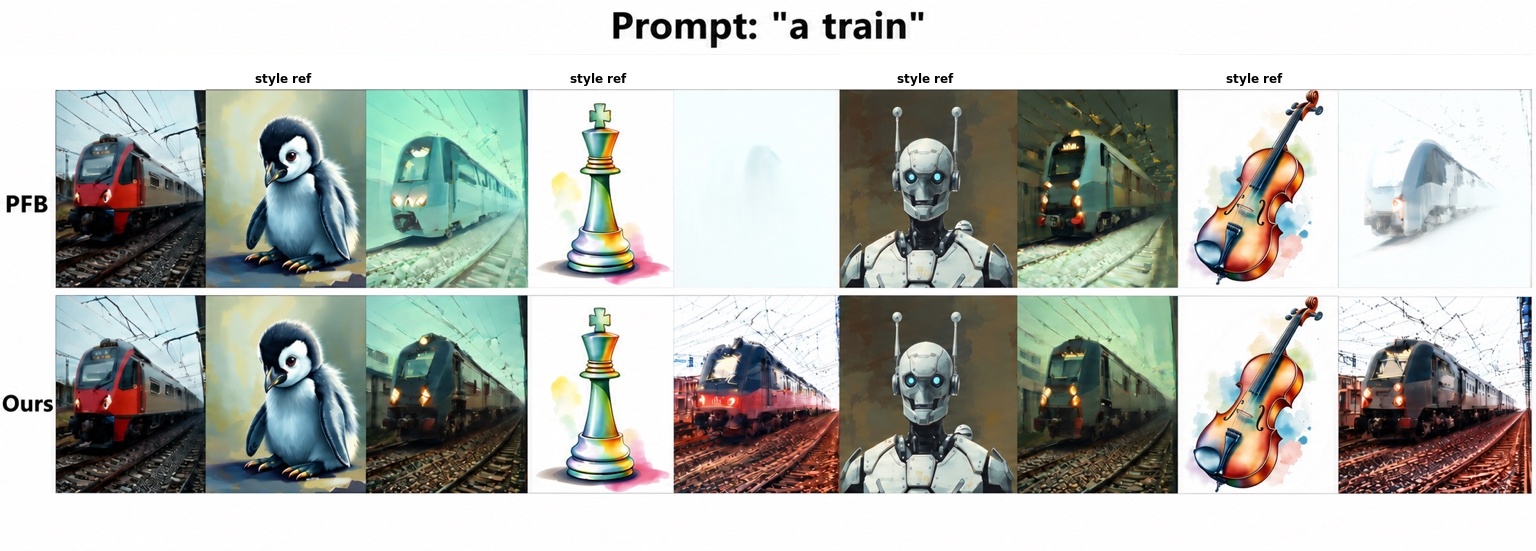}

        \captionof{figure}{Same-style references with different objects for \emph{chair}, \emph{dragon}, and \emph{train}. Projection reduces sensitivity to the reference object.}
        \label{fig:svd_sensitivity}
    \end{minipage}

    \vspace{0mm}

    \begin{minipage}{\textwidth}
        \centering
        \small
        \setlength{\tabcolsep}{3pt}
        \captionof{table}{Same-style sensitivity. Cos. and Dist. are mean pairwise cosine similarity and Euclidean distance between normalized representations. For outputs, $M_{\rm ref}$ is CLIP image similarity to the reference and $S_{\rm obj}$ is CLIP similarity to its object text; $\Delta$ is relative to clean Infinity.}
        \label{tab:svd_sensitivity}
        \begin{tabular}{lcccc}
            \toprule
            Representation
            & Cos.$\uparrow$
            & Dist.$\downarrow$
            & $\Delta M_{\rm ref}\downarrow$
            & $\Delta S_{\rm obj}\downarrow$ \\
            \midrule
            Top-1 component & 0.267 & 1.183 & -- & -- \\
            PFB edit        & 0.485 & 1.020 & -- & -- \\
            \midrule
            PFB output      & 0.837 & 0.394 & +0.0118 & +0.0095 \\
            Ours            & \textbf{0.913} & \textbf{0.248}
                            & \textbf{+0.0070} & \textbf{-0.0022} \\
            \bottomrule
        \end{tabular}
    \end{minipage}
\end{figure*}

\paragraph{\textbf{Content sensitivity of dominant SVD features.}}
\label{sec:svd_content_sensitivity}

We fix the nominal style and vary the depicted object using 13 CSD-100~\cite{nguyen2025csdvar} style groups, 53 references, and the prompts \emph{chair}, \emph{train}, and \emph{dragon}. Rank-one PFB at $R_3$ without SAC produces 159 outputs per method. Same-style references have dissimilar leading components, and PFB outputs vary with the reference object (Fig.~\ref{fig:svd_sensitivity}), showing that the dominant component remains content-sensitive.

\subsection{Method Overview}
\label{sec:overview}

\begin{figure*}[!t]
    \centering
    \makebox[\textwidth][c]{%
        \includegraphics[width=\textwidth]{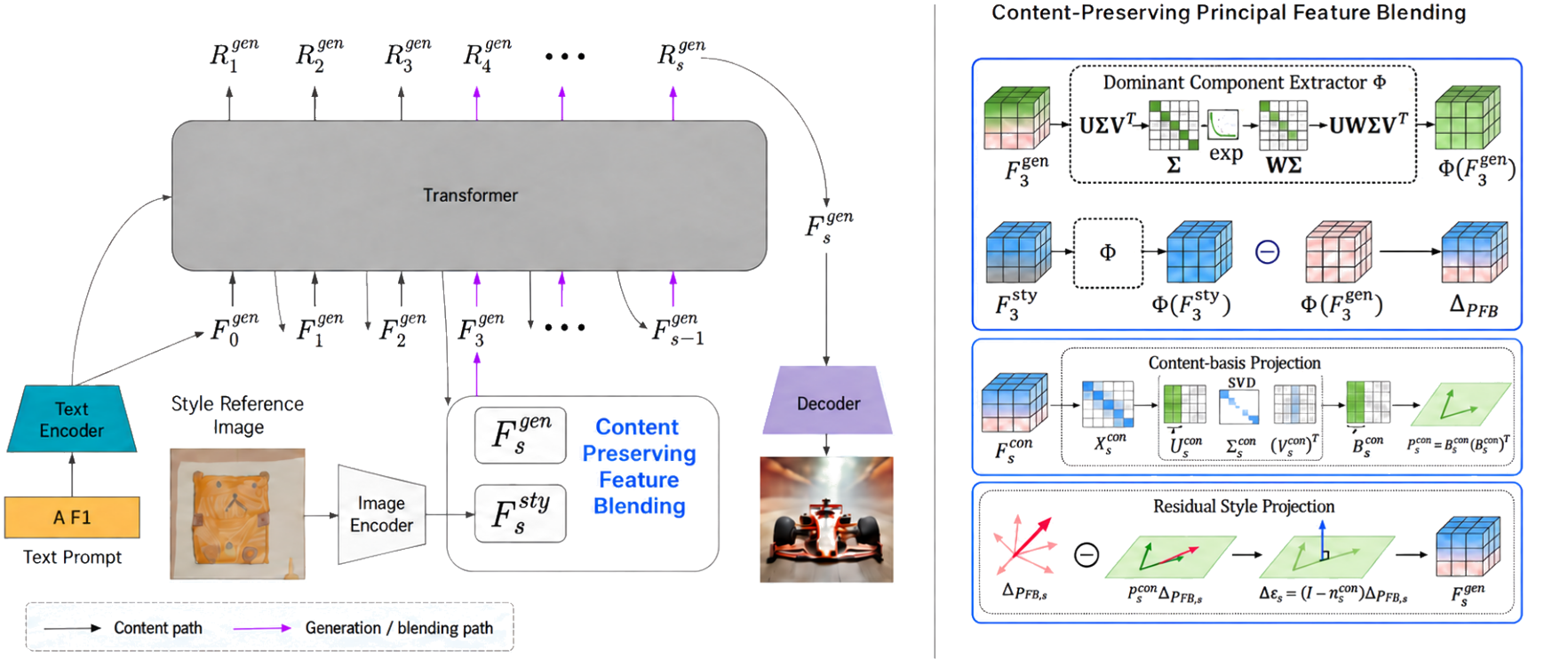}%
    }
    \caption{CoPFB overview. Black arrows show the clean content path; purple arrows show generation and blending.}
    \label{fig:cb_pfb_overview}
\end{figure*}

Given a content prompt $T^{\mathrm{con}}$ and style reference $I^{\mathrm{sty}}$, we aim to preserve prompt semantics while reproducing reference appearance using frozen Infinity~\cite{han2024infinity}, without training or learnable parameters.
As shown in Fig.~\ref{fig:cb_pfb_overview}, a clean pass provides content anchors $\{\mathbf{F}^{\mathrm{con}}_s\}_{s=1}^{S}$, an editable pass provides $\{\mathbf{F}^{\mathrm{gen}}_s\}_{s=1}^{S}$, and the visual tokenizer encodes the reference as $\{\mathbf{F}^{\mathrm{sty}}_s\}_{s=1}^{S}$.
At each selected scale, CoPFB constructs a PFB residual, removes the component aligned with a content basis estimated from the clean feature, and injects the remainder with scale-dependent strength. SAC~\cite{Lee_2026_CVPR} uses the clean stream as structural guidance during refinement.

\subsection{Principal Feature Blending}
\label{sec:pfb}

Following PFB~\cite{Lee_2026_CVPR}, we extract dominant singular components from an intermediate feature. For
$\mathbf{X}=\operatorname{mat}(\mathbf{F})\in\mathbb{R}^{C\times N}$ with
$\mathbf{X}=\mathbf{U}\boldsymbol{\Sigma}\mathbf{V}^{\top}$, we use the
exponentially weighted rank-$r$ reconstruction
\begin{equation}
    \Phi_{r,\alpha}(\mathbf{X})
    =
    \mathbf{U}_{:,1:r}
    \operatorname{diag}
    \left(
        \sigma_i e^{-\alpha(i-1)}
    \right)_{i=1}^{r}
    \mathbf{V}_{:,1:r}^{\top},
\end{equation}
where $\alpha$ attenuates higher-order components. Rather than restating the
full PFB procedure, we use its blending residual at scale $s$:
\begin{equation}
    \boldsymbol{\Delta}^{(r_s)}_{\mathrm{PFB},s}
    =
    \Phi_{r_s,\alpha}
    \left(\mathbf{X}^{\mathrm{sty}}_s\right)
    -
    \Phi_{r_s,\alpha}
    \left(\mathbf{X}^{\mathrm{gen}}_s\right),
    \label{eq:pfb_direction}
\end{equation}
where $\mathbf{X}^{\mathrm{sty}}_s=\operatorname{mat}(\mathbf{F}^{\mathrm{sty}}_s)$,
$\mathbf{X}^{\mathrm{gen}}_s=\operatorname{mat}(\mathbf{F}^{\mathrm{gen}}_s)$,
and $r_s$ is the selected style rank. Our method modifies this residual before
injection because its dominant directions may also contain reference content.

\subsection{Content-Conditioned Residual Projection}
\label{sec:content_projection}

Because singular values rank energy rather than semantic role, we treat the PFB residual as an entangled edit. CoPFB estimates dominant channel directions from the clean content stream and removes the residual component aligned with that subspace.
Let $\mathbf{X}^{\mathrm{con}}_s\in\mathbb{R}^{C\times N}$ denote the
flattened clean-content feature at scale $s$. We center it across the $N$
spatial tokens and form a channel-space basis from its leading left singular
vectors:
\begin{align}
\boldsymbol{\mu}^{\mathrm{con}}_s
&=
\frac{1}{N}\mathbf{X}^{\mathrm{con}}_s\mathbf{1}_N,
&
\widetilde{\mathbf{X}}^{\mathrm{con}}_s
&=
\mathbf{X}^{\mathrm{con}}_s
-
\boldsymbol{\mu}^{\mathrm{con}}_s\mathbf{1}_N^{\top},
\\[-1mm]
\widetilde{\mathbf{X}}^{\mathrm{con}}_s
&=
\mathbf{U}^{\mathrm{con}}_s
\boldsymbol{\Sigma}^{\mathrm{con}}_s
\left(\mathbf{V}^{\mathrm{con}}_s\right)^{\top},
&
\mathbf{B}^{\mathrm{con}}_s
&=
\left(\mathbf{U}^{\mathrm{con}}_s\right)_{:,1:r_c},
\quad
\mathbf{P}^{\mathrm{con}}_s
=
\mathbf{B}^{\mathrm{con}}_s
\left(\mathbf{B}^{\mathrm{con}}_s\right)^{\top}.
\label{eq:content_basis}
\end{align}
Centering removes spatially constant offsets, and the left singular vectors match PFB's channel space. Thus, $\mathbf{P}^{\mathrm{con}}_s$ projects onto dominant target-content directions.
Given a candidate PFB direction $\boldsymbol{\Delta}_s$, we suppress its
content-aligned component before injection:
\begin{equation}
\boldsymbol{\Delta}^{\mathrm{res}}_s
=
\left(
\mathbf{I}_C-\eta\mathbf{P}^{\mathrm{con}}_s
\right)
\boldsymbol{\Delta}_s,
\qquad \eta\in[0,1].
\label{eq:content_conditioned_update}
\end{equation}
Here, $\eta=0$ recovers PFB and $\eta=1$ removes the full content-aligned component. We call $\boldsymbol{\Delta}^{\mathrm{res}}_s$ a \emph{content-filtered residual}, because its orthogonal complement may still contain entangled information.
\subsection{Hybrid Multi-Scale Style Injection}
\label{sec:hybrid_injection}

The scale-wise analysis motivates an unprojected rank-one edit at $R_1$--$R_2$ and a content-projected rank-one edit at $R_3$--$R_5$, followed by no injection.
Let $\boldsymbol{\Delta}^{(k)}_{\mathrm{PFB},s}$ be the rank-$k$ PFB residual
from Eq.~\eqref{eq:pfb_direction}. We apply the simple update
\begin{equation}
    \widehat{\mathbf{X}}^{\mathrm{gen}}_s
    =
    \mathbf{X}^{\mathrm{gen}}_s
    +\lambda_s\boldsymbol{\Delta}_s,
    \qquad
    \lambda_s=d^{j(s)},
    \quad j(s)\in\{0,\ldots,4\},
\label{eq:hybrid_injection}
\end{equation}
where $j(s)$ is the injection order over $s\in\{1,2,3,4,5\}$ and $d=0.75$. The residual is selected by scale as
\begin{equation}
\boldsymbol{\Delta}_s
=
\begin{cases}
\boldsymbol{\Delta}^{(1)}_{\mathrm{PFB},s},
    & s\in\{1,2\},\\[1mm]
\left(\mathbf{I}_C-\eta\mathbf{P}^{\mathrm{con}}_s\right)
\boldsymbol{\Delta}^{(1)}_{\mathrm{PFB},s},
    & s\in\{3,4,5\},\\[1mm]
\mathbf{0}, & s\in\{0,6,7,8,9\}.
\end{cases}
\end{equation}
This is the released configuration: $r_c=1$, $\eta=1$, $\alpha=1$, and $\lambda_s\in\{1,0.75,...,0.75^4\}$. The mean displacement is not added separately. SAC is enabled from $R_2$ onward with unit strength.

\section{Experiments}

\subsection{Implementation Details}

We use the frozen Infinity-2B Q8$_0$ checkpoint~\cite{han2024infinity} at $512\!\times\!512$ resolution with 10 autoregressive scales. Transformer sampling uses bfloat16 autocast, while SVD and content-basis estimation use float32. We set CFG to 3.0, temperature to 0.5, top-$k$ to 900, top-$p$ to 0.97, and seed to 42 for every paired comparison. CoPFB applies rank-one PFB at zero-based scales 1--2 and projected rank-one PFB at scales 3--5 with $\alpha=1$, $r_c=1$, $\eta=1$, decay 0.75, and no mean-preservation term. SAC has unit strength and is active from scale 2 through scale 9. The visual tokenizer encodes the style reference; all model parameters remain frozen.
\subsection{Evaluation Setup}

Following FineStyle~\cite{zhang2024finestyle}, we pair 190 object-centered Parti prompts~\cite{yu2022parti} with 10 FineStyle reference images, producing 1,900 prompt-reference pairs. Each prompt specifies a subject and superclass; all methods use identical prompts, references, seeds, sampling settings, and resolution.
Let $\bar f_I$ and $\bar f_T$ be normalized CLIP-ViT-B/32 embeddings~\cite{radford2021clip}. For output $G$, reference $I^{\mathrm{sty}}$, and text $T$ concatenating the content prompt, superclass, and style descriptor, $S_{\mathrm{txt}}=\max(0,\bar f_I(G)^\top\bar f_T(T))$ and $S_{\mathrm{img}}=\max(0,\bar f_I(G)^\top\bar f_I(I^{\mathrm{sty}}))$. Their harmonic mean is
\begin{equation}
    S_{\mathrm{harmonic}}
    =
    \frac{2S_{\mathrm{txt}}S_{\mathrm{img}}}
    {S_{\mathrm{txt}}+S_{\mathrm{img}}}.
    \label{eq:harmonic_score}
\end{equation}

We report means over all pairs and use $S_{\mathrm{harmonic}}$ as a compact content-style summary. Because CLIP image similarity can respond to reference content as well as appearance, $S_{\mathrm{img}}$ is an imperfect proxy for style fidelity and should be interpreted together with qualitative results.
\begin{figure*}[!t]
    \centering
    \begin{minipage}{\textwidth}
        \centering
        \small
        \setlength{\tabcolsep}{4pt}
        \captionof{table}{Comparison on 1,900 Parti/FineStyle pairs. All methods are training-free; time is seconds per image on one NVIDIA A100 40~GB GPU, reported as mean $\pm$ uncertainty.}
        \label{tab:quantitative_results}
        \resizebox{\textwidth}{!}{%
        \begin{tabular}{lccc}
            \toprule
            \textbf{Metric}
                & \textbf{PFB+SAC~\cite{Lee_2026_CVPR}}
                & \textbf{Content Projection ($R_3$)}
                & \textbf{\shortstack{Hybrid: SVD $R_{1:2}$;\\Projection $R_{3:5}$}} \\
            \midrule
            Training-free
                & $\checkmark$ & $\checkmark$ & $\checkmark$ \\
            \midrule
            Prompt fidelity ($S_{\mathrm{txt}}$)$\uparrow$
                & 0.293
                & 0.295
                & \textbf{0.297} \\
            Style fidelity ($S_{\mathrm{img}}$)$\uparrow$
                & \textbf{0.590}
                & 0.561
                & 0.577 \\
            Harmonic score ($S_{\mathrm{harmonic}}$)$\uparrow$
                & 0.391
                & 0.387
                & \textbf{0.393} \\
            \midrule
            Inference time (seconds/image)$\downarrow$
                & $3.8\!\pm\!0.6$ & $1.9\!\pm\!0.3$ & $2.0\!\pm\!0.2$ \\
            \bottomrule
        \end{tabular}%
        }
    \end{minipage}

    \vspace{0mm}

    \begin{minipage}{\textwidth}
        \centering
        \includegraphics[width=\textwidth]{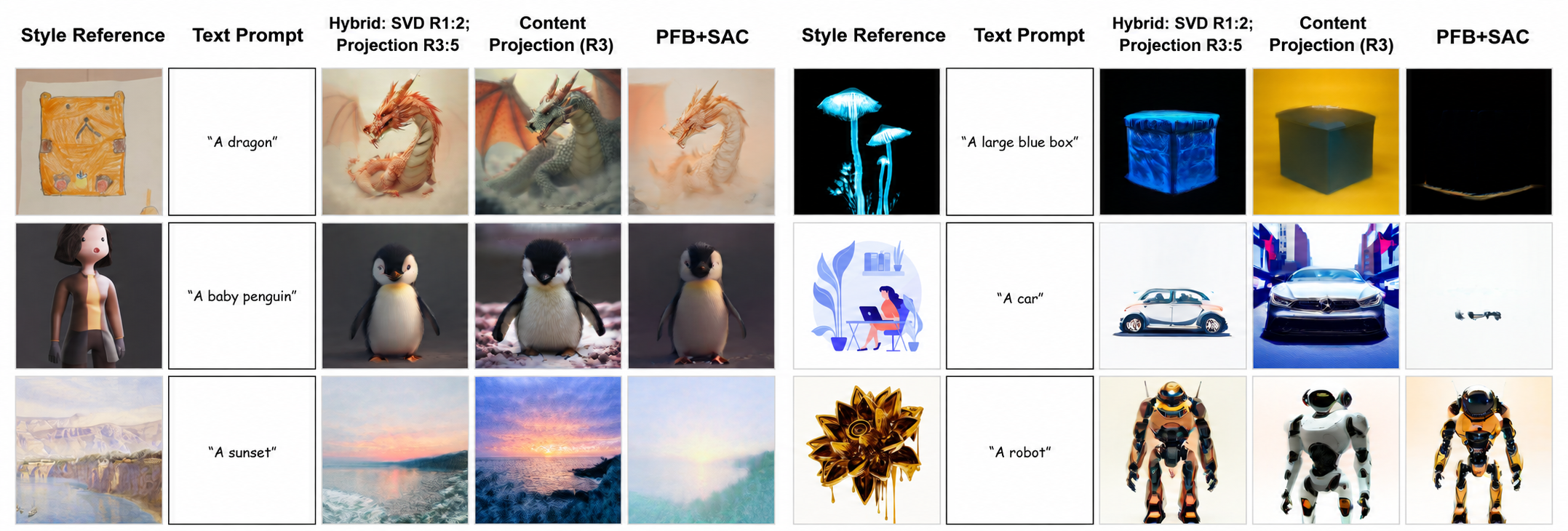}
        \captionof{figure}{Qualitative comparison on six prompt-reference pairs.}
        \label{fig:qualitative_selected}
    \end{minipage}
\end{figure*}

\subsection{Qualitative and Quantitative Comparison}

\paragraph{Qualitative.}

Figure~\ref{fig:qualitative_selected} uses pairs with a semantic mismatch between the requested and reference objects. PFB+SAC strongly transfers appearance but can suppress the requested blue box and car. Single-scale projection restores recognizable content but transfers less appearance. The hybrid configuration preserves the prompted subjects while retaining visible reference cues, including the dragon's sketch-like palette and the robot's golden ornamentation.
\paragraph{Quantitative.}

Table~\ref{tab:quantitative_results} reports dataset-level means. Relative to PFB+SAC, single-scale projection raises $S_{\mathrm{txt}}$ from 0.293 to 0.295 but lowers $S_{\mathrm{img}}$ from 0.590 to 0.561. The hybrid configuration reaches 0.297 text similarity, 0.577 image similarity, and a 0.393 harmonic score. Its 0.002 harmonic gain accompanies a 0.013 decrease in image similarity, indicating a content--style trade-off. Inference takes $2.0\!\pm\!0.2$ s/image for the hybrid configuration, $1.9\!\pm\!0.3$ for single-scale projection, and $3.8\!\pm\!0.6$ for PFB+SAC; fewer intervention scales partly explain the speedup.
\subsection{Ablation Study}

\paragraph{\textbf{Component ablation.}}
We assess how content projection (CP) interacts with Structural Attention
Correction (SAC) on the same 1,900 evaluation pairs.
\begin{table}[H]
    \centering
    \small
    \setlength{\tabcolsep}{5pt}
    \caption{Component ablation. All variants use the same PFB backbone.}
    \label{tab:component_ablation}
    \begin{tabular}{lccccc}
        \toprule
        Variant & CP & SAC & $S_{\mathrm{txt}}\uparrow$
        & $S_{\mathrm{img}}\uparrow$ & $S_{\mathrm{harmonic}}\uparrow$ \\
        \midrule
        PFB only   & --           & --           & 0.255   & 0.572   & 0.353 \\
        PFB+SAC    & --           & $\checkmark$ & 0.293          & \textbf{0.590} & \textbf{0.391} \\
        PFB+CP     & $\checkmark$ & --           & 0.284          & 0.563          & 0.378 \\
        PFB+CP+SAC & $\checkmark$ & $\checkmark$ & \textbf{0.295} & 0.561          & 0.387 \\
        \bottomrule
    \end{tabular}
\end{table}

For the single-scale $R_3$ configuration, CP without SAC obtains a 0.378 harmonic score. Adding SAC raises $S_{\mathrm{txt}}$ from 0.284 to 0.295 but produces a lower harmonic score than PFB+SAC because $S_{\mathrm{img}}$ decreases. This motivates the hybrid schedule.
\paragraph{\textbf{Content-basis rank.}}
We vary $r_c$ while keeping all other settings fixed.
\begin{table}[H]
    \centering
    \small
    \setlength{\tabcolsep}{7pt}
    \caption{Effect of content-basis rank across rank-1, rank-2 and rank-4.}
    \label{tab:content_rank_ablation}
    \begin{tabular}{cccc}
        \toprule
        $r_c$ & $S_{\mathrm{txt}}\uparrow$ & $S_{\mathrm{img}}\uparrow$
        & $S_{\mathrm{harmonic}}\uparrow$ \\
        \midrule
        1 & \textbf{0.295} & \textbf{0.561} & \textbf{0.387} \\
        2 & 0.288 & 0.554 & 0.379 \\
        4 & 0.281 & 0.547
          & 0.371 \\
        \bottomrule
    \end{tabular}
\end{table}

Scores decrease as $r_c$ grows, indicating that a larger protected subspace also removes style-relevant directions. We therefore use $r_c=1$.

\enlargethispage{2\baselineskip}
\vspace{-2mm}
\section{Conclusion and Future Work}

We presented CoPFB, a training-free method for content-preserving style personalization in visual autoregressive models. CoPFB projects the PFB residual away from a low-rank content basis estimated from the clean generation stream, mitigating content leakage from style references. Combined with a hybrid schedule over the first five scales, it improves the content-style trade-off without model modification or learnable parameters. On FineStyle with 1,900 prompt-style pairs, CoPFB achieves the strongest prompt-fidelity and harmonic scores while maintaining competitive style similarity.

Future work will investigate adaptive content-basis ranks and injection strengths that respond to the prompt, reference style, and generation scale. We also plan to evaluate richer content and style metrics, extend the method to additional autoregressive architectures, and study whether spatially localized projections can preserve fine object structure while transferring more detailed textures.

\begin{credits}
\subsubsection{\ackname} 
This research is funded by Vietnam National University - Ho Chi Minh City (VNU-HCM) under Grant Number B2026-18-17.
\end{credits}
%
%
%
\bibliographystyle{splncs04}
\bibliography{references}

\end{document}